\documentclass[extended-abstract]{ccn}

\title{Vision Models Capture Complementary Representations in Children That Converge in Adults}

\author{%
  Domenic Bersch \affmark{1,2} \And
  Timothy Schaumlöffel \affmark{1,2} \And
  Michela Proietti \affmark{1,2} \AND
  Antonia Franaszek-Traczewska \affmark{3} \And
  Hannah Elisabeth Zwad \affmark{3} \And
  Siying Xie \affmark{3} \AND
  Marlena Baldauf \affmark{4} \And
  Teresa Sylvester \affmark{3} \And
  Christina Maria Schätz \affmark{5} \And
  Bert Turtleton \affmark{3} \AND
  Moritz Köster \affmark{4} \And
  Stefanie Höhl \affmark{5} \And
  Radoslaw Cichy* \affmark{3} \And
  Gemma Roig* \affmark{1,2}
}
\affiliation{1}{Department of Computer Science, Goethe University Frankfurt, Frankfurt am Main, Germany}
\affiliation{2}{The Hessian Center for Artificial Intelligence (hessian.AI), Darmstadt, Germany}
\affiliation{3}{Department of Education and Psychology, Freie Universität Berlin, Berlin, Germany}
\affiliation{4}{Institute of Psychology, Developmental Cognitive Psychology, University of Regensburg, Regensburg, Germany}
\affiliation{5}{Department of Developmental and Educational Psychology, University of Vienna, Vienna, Austria}
\emails{bersch@em.uni-frankfurt.de, schaumloeffel@em.uni-frankfurt.de, m.proietti@em.uni-frankfurt.de, antonia.franaszek-traczewska@fu-berlin.de, h.zwad@fu-berlin.de, siying.xie@outlook.com, Marlena.Baldauf@psychologie.uni-regensburg.de, teresa.sylvester@fu-berlin.de, christina.maria.schaetz@univie.ac.at, bert.turtleton@gmail.com, Moritz.Koester@psychologie.uni-regensburg.de, stefanie.hoehl@univie.ac.at, rmcichy@zedat.fu-berlin.de, roignoguera@em.uni-frankfurt.de}

\begin{document}

\maketitle

\begin{abstract}
How the relationship between image-text aligned and vision-only models changes across development remains unknown. We applied time-resolved representational similarity analysis to EEG data from 8-year-olds, 12-year-olds, and adults. CLIP showed significant model-brain alignment already in children, whereas vision-only models aligned more strongly later in development. Residual analyses revealed complementary structure, with CLIP retaining alignment beyond an architecture-matched vision-only model from age 8. Much of the model-brain alignment reflected semantic structure, with EEG representations of category and animacy strengthening toward adulthood, while alignment with low-level image properties remained weak across age. Together, these findings suggest that image-text aligned and vision-only models capture complementary aspects of developing visual representations and may provide useful constraints for AI systems that aim to learn and adapt over time.
\end{abstract}


\section{Introduction}
Deep neural networks (DNNs) are useful computational probes of visual representations because different training objectives can capture different aspects of representational geometry. In adults, representational similarity analysis (RSA) has shown that DNNs capture key aspects of visual representations measured with EEG, MEG, and fMRI \citep{CichyKhoslaPantazisEtAl2016, GIFFORD2022119754, Kriegeskorte}. In particular, adult work suggests that image-text aligned models capture aspects of visual representations not captured by architecture-matched vision-only models \citep{WangKayNaselarisTarrWehbe2023, ChenLiuWangEtAl2025}. Developmental comparisons between DNNs and EEG remain limited \citep{XIE20225422}, and it remains unclear whether the relationship between image-text aligned and vision-only models changes across development. Here, we address this question using time-resolved RSA on EEG data from 8-year-olds, 12-year-olds, and adults.

\section{Methods}
\paragraph{Dataset and preprocessing.}
We analyzed EEG data from 8-year-olds, 12-year-olds, and adults viewing centered $300 \times 300$ pixel images from four categories: faces, toys, scenes, and body parts. Analyses focused on 48 familiar images (12 per category). Data were resampled to 250 Hz, baseline-corrected using a 500\,ms prestimulus interval, and analyzed from $-100$ to 600\,ms relative to stimulus onset across 63 EEG channels (176 timepoints). The sample included 8-year-olds ($n=16$), 12-year-olds ($n=11$), and adults ($n=20$).

\vspace{-1em}

\paragraph{Models.} We compared image-text aligned and vision-only models. The primary models were CLIP-ViT-B/16 \citep{radford2021learningtransferablevisualmodels}, ImageNet-supervised ViT-B/16 \citep{dosovitskiy2021imageworth16x16words}, DINOv1-ViT-B/16 \citep{caron2021emergingpropertiesselfsupervisedvision}, DINOv2 \citep{oquab2024dinov2learningrobustvisual}, and ResNet-50 \citep{he2015deepresiduallearningimage}. We additionally compared CLIP and SimCLR ViT-B/16 models trained on YFCC15M for 25 epochs, providing a control matched for architecture, training data, and training duration \citep{mu2022slip}. We extracted layerwise features and computed RDMs as $1-\mathrm{Pearson}$ correlation between feature vectors.

\vspace{-1em}

\begin{figure}[t!]
  \begin{center}
    \includegraphics[width=\linewidth]{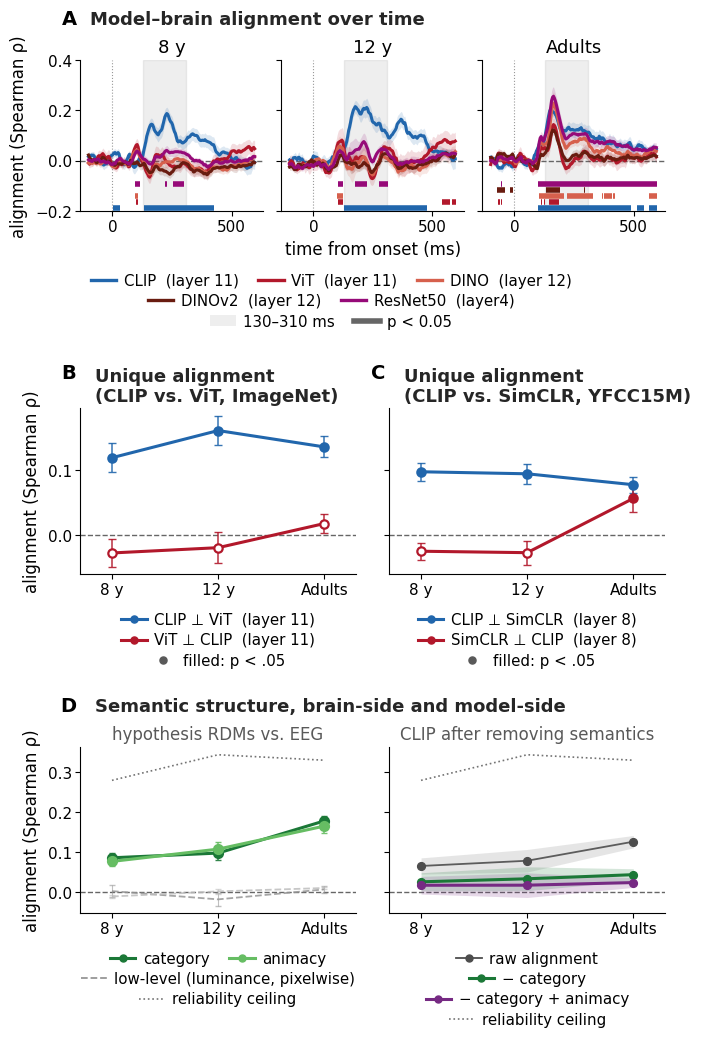}
  \end{center}
    \caption{\textbf{Model-brain alignment across development.}
    \textbf{(A)} Time-resolved RSA between five vision models and EEG for 8-year-olds, 12-year-olds, and adults. Gray bands mark the 130-310\,ms analysis window, shading indicates $\pm$SEM, and bars mark significant timepoints from uncorrected stimulus-shuffle tests.
    \textbf{(B)} Residual alignment for CLIP and ViT-B/16 after removing shared model structure, evaluated at a layer selected independently of EEG from model-model RDM similarity.
    \textbf{(C)} Corresponding analysis for CLIP and SimCLR ViT-B/16 models trained on YFCC15M for 25 epochs.
    \textbf{(D)} Left, EEG alignment with category, animacy, luminance, and pixel-level hypothesis RDMs. Right, CLIP alignment after removing category and animacy structure, averaged across layers. Shading indicates SD across layers and dotted lines the reliability ceiling.}
  \label{fig:exp1}
  \vspace{-1em}
\end{figure}

\paragraph{Representational similarity analysis.} At each timepoint, we computed $48 \times 48$ EEG representational dissimilarity matrices (RDMs) as $1-\mathrm{Pearson}$ correlation across channels and compared them with model RDMs using Spearman correlation. To isolate complementary model structure, we regressed one model RDM from another and correlated the residual with the EEG RDM. Summary statistics were computed in the 130-310\,ms window, spanning the N170 and P2 components \citep{bentin1996electrophysiological, rossion2012n170}. Statistical significance was assessed using stimulus-shuffle permutation tests (2,000 permutations for Figure~\ref{fig:exp1}A and D left; 5,000 for B--C and developmental growth contrasts).

\noindent

\section{Results}
Across 8-year-olds ($n=16$), 12-year-olds ($n=11$), and adults ($n=20$), we first examined time-resolved model-EEG correspondence (Figure~\ref{fig:exp1}A). CLIP showed significant alignment already in 8-year-olds, whereas vision-only models aligned later and showed robust alignment in adults. Summary analyses used the 130-310\,ms window and stimulus-shuffle permutation tests.

We next asked whether CLIP captured neural structure beyond an architecture-matched vision-only model. For residual analyses, layers were selected independently of EEG based on where the model RDMs were least correlated. At the maximally divergent CLIP-ViT layer, CLIP retained significant alignment beyond ViT already at age 8 ($r=.12$, $p<.001$; Figure~\ref{fig:exp1}B), indicating a CLIP-specific component of neural representational geometry present from childhood. The reverse residual is shown for comparison.

We then examined CLIP and SimCLR ViT-B/16 models matched for architecture and training data and trained on YFCC15M for 25 epochs (Figure~\ref{fig:exp1}C). In an exploratory analysis, CLIP again showed alignment beyond SimCLR in children at an intermediate layer. Overall alignment was lower for these shorter-trained checkpoints, so only within-pair effects and their developmental changes were interpreted.

Finally, we examined EEG representational structure directly using hypothesis RDMs derived from the stimulus set. These encoded semantic category membership and animacy, alongside luminance and pixel-level controls. EEG alignment with both category and animacy approximately doubled from childhood to adulthood ($p<.001$ for category growth; paired stimulus-shuffle test), whereas alignment with low-level image properties remained weak across age (Figure~\ref{fig:exp1}D). Removing category and animacy structure substantially reduced CLIP-EEG alignment across age, indicating that much of the brain-relevant structure captured by CLIP is semantic.

\section{Discussion}
Our results show that the relationship between vision models and brain representations changes systematically from childhood to adulthood. CLIP captures brain-relevant structure already in 8-year-olds that is not captured by an architecture-matched vision-only model, while the correspondence of vision-only models with brain representations becomes stronger toward adulthood. Rather than one model family simply providing a better account of visual representations, the models therefore capture complementary aspects whose relevance changes with age.

The developmental increase in model-brain alignment was primarily semantic. Category and animacy structure were already present in children's EEG representations but strengthened substantially toward adulthood, whereas low-level luminance and pixel similarity showed no comparable developmental change. The matched CLIP-SimCLR comparison further suggests that architecture and training data alone do not explain the early CLIP alignment, although differences in the broader learning objective prevent us from isolating language supervision specifically. Together, these results suggest an early representational component captured particularly well by CLIP alongside a strengthening of semantic structure across development.

More broadly, developmental model-brain comparisons may help identify computational mechanisms underlying how visual representations change with experience. Future work will extend these analyses from infancy to adulthood and use model ablations and feature steering to identify components driving age-specific alignment. Such developmental constraints may ultimately inform AI systems that learn and adapt over time rather than only reproduce adult-like representations.

\section{Acknowledgements and Disclosure}
This work was funded by the BMFTR neuroAI grant GROWUPAI (R.M.C., G.R.), the German Research Council (DFG) grants (CI 241/1-3, CI 241/1-7, INST 272/297-2) (R.M.C.), DFG project 5368 (ARENA) and RO 6458/5-1 (G.R.), and a European Research Council (ERC) consolidator grant (ERC-CoG-2024101123101) (R.M.C.).\\
\\
Large language models were used for language editing, grammar refinement, and minor programming assistance for plotting code. All scientific claims, analyses, interpretations, and final manuscript content were reviewed and approved by the authors.

\printbibliography

\end{document}